%% file: main.tex
\documentclass{article}

\usepackage{microtype}
\usepackage{graphicx}
\usepackage{subcaption}
\usepackage{booktabs}
\usepackage{algorithm}
\usepackage{algorithmic}
\usepackage{amsmath}
\usepackage{amssymb}
\usepackage{mathtools}
\usepackage{amsthm}
\usepackage{enumitem}
\usepackage{xcolor}
\usepackage[most]{tcolorbox}
\usepackage{placeins}
\usepackage{url}
\usepackage[capitalize,noabbrev]{cleveref}

\usepackage[accepted]{icml2026}
\makeatletter
\renewcommand{\Notice@String}{%
  \textit{Proceedings of the
$\mathit{43}^{rd}$ International Conference on Machine Learning Workshop} on AI as a Tool for Mathematics, Computer Science, and Machine Learning,
Seoul, South Korea. PMLR 306, 2026.
Copyright 2026 by the author(s).}%
\makeatother

\newcommand{\lite}{\textsc{CKM-Lite}}
\newcommand{\full}{\textsc{CKM-Full}}
\newcommand{\batch}{\textsc{Batch}}
\newcommand{\ckm}{\textsc{CKM}}

\icmltitlerunning{Continuous Knowledge Metabolism}

\begin{document}

\twocolumn[
  \icmltitle{Continuous Knowledge Metabolism:\\
             Generating Scientific Hypotheses from Evolving Literature}

  \begin{icmlauthorlist}

    \icmlauthor{Jinkai Tao}{inst1,inst2,inst3}
    \icmlauthor{Yubo Wang}{inst4}
    \icmlauthor{Xiaoyu Liu}{inst1}
    \icmlauthor{Menglin Yang}{inst3}

  \end{icmlauthorlist}
  
  \icmlaffiliation{inst1}{TsingyuAI, Beijing, China}
  \icmlaffiliation{inst2}{School of Information, Central University of Finance and Economics, Beijing, China}
  \icmlaffiliation{inst3}{Thrust of Artificial Intelligence, The Hong
Kong University of Science and Technology (Guangzhou), Guangzhou, China}
  \icmlaffiliation{inst4}{School of Automation, Beijing Institute of Technology, Beijing, China}

  \icmlcorrespondingauthor{Jinkai Tao}{jinkaitao.comm@gmail.com}
  \icmlcorrespondingauthor{Menglin Yang}{menglin.yang@outlook.com}

  \icmlkeywords{AI-assisted research, scientific hypothesis generation, literature workflow, large language models}

  \vskip 0.3in
]

\printAffiliationsAndNotice{}

\input{sections/00abstract}
\input{sections/01introduction}
\FloatBarrier
\input{sections/02method}
\FloatBarrier
\input{sections/03experiments}
\FloatBarrier
\input{sections/04failures}
\FloatBarrier
\input{sections/05conclusion}

\input{sections/06impact}

\bibliography{main}
\bibliographystyle{icml2026}

\newpage
\appendix
\onecolumn

\input{sections/07appendix}

\end{document}

%% file: sections/00abstract.tex
\begin{abstract}
    Identifying promising research directions in fast-moving subareas is one of the most cognitively expensive tasks in modern AI research. Existing LLM-driven scientific-discovery systems are typically limited to one-shot prompting on static literature snapshots and validated only against contemporary judges (human reviewers, agent peer review, wet-lab assays, or self-evaluation), leaving open whether they anticipate future trends.
    
    We present \emph{Continuous Knowledge Metabolism} (\ckm{}), an AI workflow for hypothesis generation with three key capabilities:
    (i) \emph{continuous literature metabolism} via sliding windows that maintain an evolving knowledge state;
    (ii) \emph{predictive evaluation}, grading hypotheses against papers published \emph{after} the generation window; and
    (iii) \emph{practitioner-grade failure detection} that diagnoses workflow failure modes from its outputs.
    
    On a 50-topic ML benchmark, \lite{} produces at least one validated hypothesis on $72\%$ of topics ($36/50$), more than doubling a one-shot baseline ($30\%$) at $\sim$\$3 per topic and $91\%$ lower token cost. Validated hypotheses precede their matched papers by a mean of $404$ days ($n{=}55$ hits across $36$ topics; median $399$, range $66$--$757$).
    
    Broadly, predictive validation against future literature offers a falsifiable, low-cost alternative to contemporary-judge protocols and can be applied wherever a corpus has dated publication records.
    \end{abstract}
    

%% file: sections/01introduction.tex
\section{Introduction}
\label{sec:intro}

Picking the next research project is one of the most expensive cognitive acts in modern AI research.
Before committing a quarter to a new direction, a researcher needs to know what is under-explored right now: not what last year's top venues said, but what the last six months' papers point at.
Doing this by hand means reading 30--60 papers, maintaining cross-session notes, tracking what changed since last quarter, and drafting actionable research questions.
A careful pass takes 12--20 hours per topic; a researcher who wants to scope many directions in parallel simply runs out of time.

LLMs have started to compress this loop.
A small but rapidly growing literature on LLM-driven scientific discovery now covers idea generation~\citep{wang-etal-2024-scimon, baek2025researchagent}, hypothesis generation~\citep{zhou-etal-2024-hypothesis, gottweis2025towards}, and end-to-end paper writing~\citep{lu2025aiscientist, lyu2026evoscientist} (Appendix~\ref{app:related}).
Across this literature, validation follows a consistent template, in which a contemporary judge (a human reviewer~\citep{si2025can}, an agent peer-review loop~\citep{weng2025cycleresearcher}, a tournament~\citep{hu2025nova}, a wet-lab assay~\citep{gottweis2025towards}, or self-evaluation~\citep{lu2025aiscientist}) decides whether the output looks good.
What no system in the list tests is whether subsequent literature actually pursued the proposed direction, even though that is arguably the only criterion that matters for a system claiming to anticipate where a field is heading.

We present \emph{Continuous Knowledge Metabolism} (\ckm{}), an AI workflow that turns moving literature into ranked, testable hypotheses and grades each one against papers published \emph{after} the generation window.
On a 50-topic ML benchmark, \lite{} produces at least one validated hypothesis on $72\%$ of topics ($36/50$), more than doubling a one-shot baseline ($30\%$) at $91\%$ lower token cost; the workflow runs at $\sim\!\$3$ per topic on a laptop.
We further isolate three regime boundaries (\S\ref{sec:failures}), each paired with a user-computable detection signal that turns the trade-off from a hidden failure into a knob.

%% file: sections/02method.tex
\section{Method: The CKM Workflow}
\label{sec:method}

\begin{figure*}[!t]
    \centering
    \includegraphics[width=0.95\textwidth]{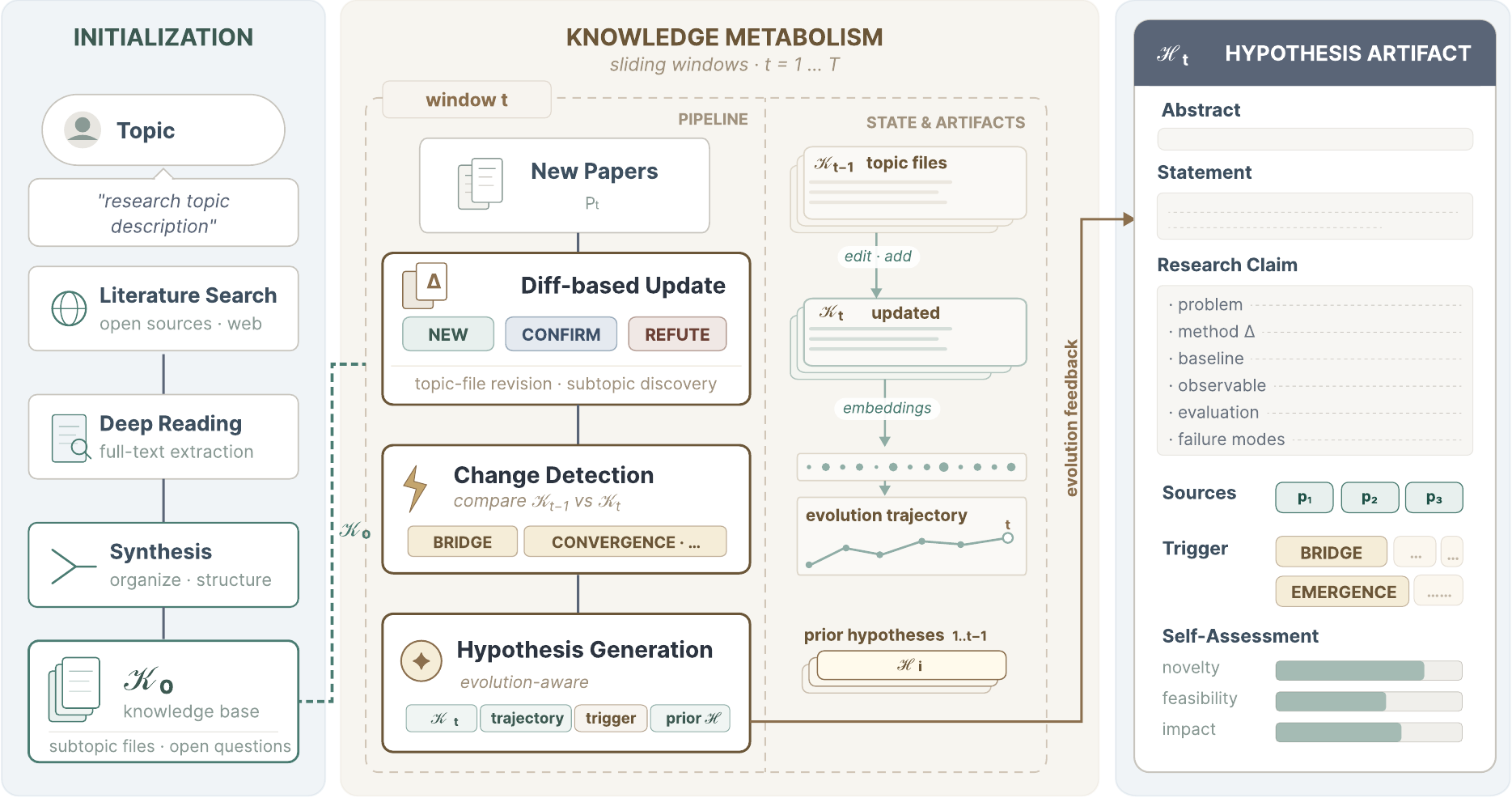}
    \vspace{-2mm}
    \caption{The \ckm{} workflow. \emph{Initialization} reads historical papers into a structured knowledge state $\mathcal{K}_0$. Each evolution window $t$ updates $\mathcal{K}_{t-1}\!\to\!\mathcal{K}_t$ with new papers and emits hypotheses $\mathcal{H}_t$. \lite{} implements the core cycle; \full{} adds an explicit change-detection step (dashed) that we ablate in \S\ref{sec:failures}.}
    \label{fig:framework}
\end{figure*}

We name the workflow \emph{Continuous Knowledge Metabolism} (\ckm{}): it absorbs literature incrementally, integrates each finding contrastively with the existing state, and maintains a substrate that anticipates field directions rather than summarizing past work.  
\ckm{} runs in three stages (Figure~\ref{fig:framework}) as an instance of \emph{tool-augmented reasoning}: deterministic tools (\textsc{arxiv\_search}, \textsc{openalex\_search}) plus an embedding selector, restricted to arXiv for consistency. The variant \lite{} implements the cycle; \full{} adds a change-detection probe (\S\ref{sec:failures}).

\paragraph{Stage 1: Initialization.}
We retrieve up to $48$ pre-window papers, prompt the LLM to extract structured per-paper notes (core methods, key findings, datasets, open questions), then cluster these into Markdown \emph{topic files} partitioned by sub-theme; the union is the initial knowledge state $\mathcal{K}_0$. The sub-theme partition is itself LLM-induced from the abstracts, not pre-specified, so $\mathcal{K}_0$ adapts to each topic's actual sub-structure rather than imposing a fixed taxonomy.

\paragraph{Stage 2: Window update.}
The evolution period is sliced into sliding windows of width $\Delta t$. The update $\mathcal{K}_t = \mathrm{Update}(\mathcal{K}_{t-1}, \mathcal{P}_t)$ folds in-window papers $\mathcal{P}_t$ into $\mathcal{K}_{t-1}$, appending new claims, strengthening confirming evidence, and surfacing contradictions; the contrast against $\mathcal{K}_{t-1}$ is what distinguishes this from a rolling summary. Mechanically, $\mathrm{Update}$ is a single LLM call that edits the topic files in place: new sub-themes are added when warranted, new evidence is attached to confirmed claims, and contradictions are tagged explicitly so they persist across windows rather than being averaged away. The LLM never re-reads the historical corpus, keeping the per-topic budget at $\sim$0.5M tokens (\S\ref{sec:eval}).

\paragraph{Stage 3: Hypothesis generation.}
The generator conditions on three contexts: $\mathcal{K}_t$, a running summary $\mathcal{S}_{1:t-1}$ of cross-window changes, and prior hypotheses $\mathcal{H}_{1:t-1}$ (so the model proposes emerging directions and avoids re-proposing); $\mathcal{S}$ lets the model distinguish a fresh trend from a long-standing direction. Each hypothesis is a structured JSON object (one-sentence claim; research plan with problem, method delta, baseline, expected observable, and evaluation; source citations; self-assessed novelty/feasibility/impact), making downstream verification (\S\ref{sec:eval}) tractable. Generation is throttled to $\sim$3 hypotheses per window under structured-output decoding, so per-topic yield is schedule-determined, not temperature-determined.

\begin{algorithm}[!t]
\caption{\ckm{} workflow over one topic.}
\label{alg:ckm}
\begin{algorithmic}[1]
\REQUIRE topic $q$, init range $[t_\text{init}, t_0]$, window width $\Delta t$, horizon $T$
\STATE $\mathcal{K}_0 \leftarrow \mathrm{Init}(q, \mathcal{P}^{\text{init}})$;\quad $\mathcal{S}_0 \leftarrow \emptyset$;\quad $\mathcal{H} \leftarrow \emptyset$
\FOR{$t = 1, \dots, T$}
  \STATE $\mathcal{P}_t \leftarrow$ fetch papers in window $t$
  \STATE $\mathcal{K}_t \leftarrow \mathrm{Update}(\mathcal{K}_{t-1}, \mathcal{P}_t)$
  \STATE $\mathcal{H}_t \leftarrow \mathrm{Gen}(q, \mathcal{K}_t, \mathcal{S}_{1:t-1}, \mathcal{H}_{1:t-1})$
  \STATE $\mathcal{S}_t \leftarrow$ summary of $\mathcal{K}_{t-1}\!\to\!\mathcal{K}_t$;\quad $\mathcal{H} \leftarrow \mathcal{H} \cup \mathcal{H}_t$
\ENDFOR
\STATE \textbf{return} $\mathcal{H}$, $\mathcal{K}_T$
\end{algorithmic}
\end{algorithm}

\begin{table*}[!t]
  \centering
  \footnotesize
  \setlength{\tabcolsep}{8pt}
  \renewcommand{\arraystretch}{1.05}
  \begin{tabular}{@{}lcccccccc@{}}
  \toprule
  & \multicolumn{4}{c}{\textbf{Predictive performance}} & \multicolumn{4}{c}{\textbf{Per-hypothesis novelty (1--10)}} \\
  \cmidrule(lr){2-5} \cmidrule(lr){6-9}
  \textbf{System} & \textbf{Yield} & \textbf{Hit\%} & \textbf{Coverage} & \textbf{Tokens} & \textbf{D1 Orig.} & \textbf{D2 Cross-f.} & \textbf{D3 Gap} & \textbf{D4 Falsif.} \\
  \midrule
  \lite{} \textit{(this work)} & \textbf{17.3} & \textbf{5.8} & \textbf{36/50} & \textbf{0.51M} & 4.20          & 5.78          & 6.21          & \textbf{7.86} \\
  \batch{}                     & 13.7          & 3.0          & 15/50          & 5.93M          & 4.77          & 6.22          & \textbf{6.75} & 7.71 \\
  \full{}                      & 17.8          & 1.4          & 11/50          & 5.87M          & \textbf{5.64} & \textbf{6.62} & 6.59          & 7.84 \\
  \bottomrule
  \end{tabular}
  \vspace{2pt}
  \caption{Main results across 50 ML topics. \emph{Left:} predictive performance (coverage = topics with $\geq$1 hit). \emph{Right:} per-hypothesis novelty, scored by an independent LLM judge on a 1--10 scale across four dimensions (full rubric in Appendix~\ref{app:novelty}). Bold = best per column; all systems run 53--67 min per topic.}
  \label{tab:main}
  \end{table*}

%% file: sections/03experiments.tex
\section{Experiments}
\label{sec:eval}

\paragraph{Benchmark.}
We selected 50 ML research topics spanning eight categories (full list in Appendix~\ref{app:topics}), ranging from slow-moving areas (machine translation, information retrieval) to fast-moving LLM topics (chain-of-thought, RLHF, tool-using agents) and applied areas (drug discovery, clinical NLP).
Unlike IdeaBench~\citep{guo2025ideabench} and LiveIdeaBench~\citep{ruan2026evaluating}, which judge ideas with contemporary scorers, we anchor evaluation in dated future literature, forcing a temporal split.
Each topic is split into three temporal phases: \emph{initialization} ($\leq$48 papers from 2019--2024 build the baseline state $\mathcal{K}_0$), \emph{evolution} (six 2-month sliding windows over 2024--2025, $\leq$96 papers per window, generate hypotheses), and \emph{validation} (papers from 2025--2027 serve as future ground truth).
A generated hypothesis is judged a \emph{hit} if at least one validation paper aligns with it at score $\geq 6.0$ on a 1--10 LLM-judge scale; judging uses a two-stage cross-provider protocol (GPT-4o-mini pre-filter then GPT-4o re-judgment~\citep{openai2024gpt4ocard}, both different from the Gemini-2.5-Flash generator~\citep{comanici2025gemini25pushingfrontier}; full safeguards in Appendix~\ref{app:safeguards}).

\paragraph{Baseline.}
The natural baseline is \batch{}: dump all in-window papers into a single prompt and ask for hypotheses: what most LLM-using researchers run today.
We additionally include \full{}, an instrumented \lite{} variant with a change-detection probe (\S\ref{sec:failures}).

\paragraph{Result.}
\lite{} produces more, better-aligned hypotheses than \batch{} at $\sim$11$\times$ lower token cost; \full{} trades coverage for per-hypothesis novelty (Table~\ref{tab:main}).
Per topic across six evolution windows, \lite{} yields $17.3$ hypotheses with a $5.8\%$ hit rate, against $13.7$ and $3.0\%$ for \batch{}; the gap is significant under a Wilcoxon paired test on per-topic hit rates over all $50$ topics ($p{=}0.006$, effect size $r=0.44$, $n_{\mathrm{eff}}=38$ after dropping ties), with \lite{} winning $29$ topics, losing $9$, and tying $12$.
Coverage is $36/50 = 72\%$ for \lite{} versus $15/50 = 30\%$ for \batch{}; per-topic generation tokens drop from $5.93$M to $0.51$M ($91.4\%$ reduction); across \lite{}'s $64$ hit papers, the mean generation-to-publication lead is $\sim$$404$ days (median $399$, range $66$--$757$).
The two halves of Table~\ref{tab:main} cover orthogonal correctness axes: predictive alignment (left) and per-hypothesis novelty against pre-window literature (right). \full{}'s change-detection pipeline produces the highest novelty (D1 originality, D2 cross-field synthesis), which \lite{} trades for the broader-tail hit-rate distribution---a high-novelty/low-hit hypothesis may lead the field rather than fail it (Lite vs Full ablation: F1, \S\ref{sec:failures}).

\begin{figure*}[t!]
    \centering
    \includegraphics[width=0.92\textwidth]{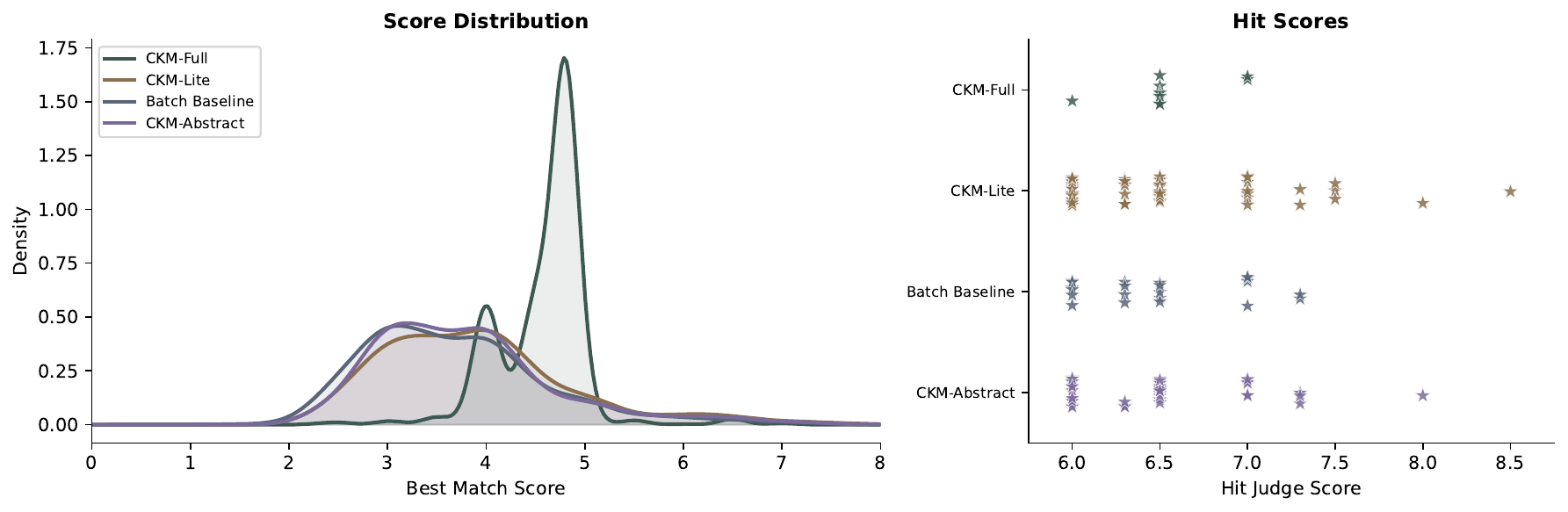}
    \vspace{-2mm}
    \caption{The quality-coverage trade-off behind F1, visualized. \emph{Left:} density of best-match alignment scores across all hypotheses for \full{} (instrumented), \lite{} (this work), and \batch{}. \full{} concentrates in the 4--5 band (high floor, narrow ceiling); \lite{} has a broader distribution with a thicker right tail crossing the hit threshold of $6.0$. \emph{Right:} individual hit scores for each system. \lite{}'s coverage advantage comes from the broad-tail population, not from a small number of high-scoring outliers.}
    \label{fig:density}
\end{figure*}

\paragraph{Validity of the result.}
Three guards against optimism: post-training-cutoff validation papers (no memorization), cross-provider generation/judging (no self-preference), and re-judging hits by a stronger judge at a higher threshold (full safeguards in Appendix~\ref{app:safeguards}).
A hit at threshold $\geq 6.0$ requires the judge to identify overlapping problem framing, methodological direction, and predicted observable between hypothesis and future paper, not surface-level keyword overlap; the threshold sits at the inflection in the score distribution (Figure~\ref{fig:density}), separating substantive from incidental matches.
The workflow itself is model-agnostic: substituting open-source generators or judges only requires changing two configuration values, not code (Appendix~\ref{app:repro}).

\begin{figure}[!b]
\begin{tcolorbox}[
  colback=white, colframe=black!35, boxrule=0.5pt, arc=1.5mm,
  left=4pt, right=4pt, top=4pt, bottom=4pt,
  title={\scriptsize\bfseries RLHF Stability \hfill Trigger: \textsc{Convergence} $\cdot$ Generated Nov 2024 $\cdot$ \textcolor{green!50!black}{Hit $\checkmark$ \textbf{460-day lead}}},
  fonttitle=\scriptsize, coltitle=black, colbacktitle=gray!10,
]
\begin{tcolorbox}[
  colback=teal!4, colframe=teal!50, boxrule=0.3pt, arc=1mm,
  left=5pt, right=5pt, top=4pt, bottom=4pt,
]
\footnotesize\itshape
``A novel RLHF framework will integrate adaptive entropy regularization (H-DPO/SEE-DPO style) with theoretically grounded KL-regularization and dynamic uncertainty-aware policy optimization, demonstrating enhanced stability and reduced reward hacking compared to standard DPO or PPO-based RLHF.''
\end{tcolorbox}

\vspace{2pt}
\footnotesize
\textbf{Matched (Feb 2026)} $\rightarrow$ \emph{SAFE: Stable Alignment Finetuning with Entropy-Aware Predictive Control for RLHF} (arXiv:2602.04651). The matched paper's title encodes the same three components named in the hypothesis: entropy-aware control, predictive/uncertainty-aware optimization, and RLHF stability.
\end{tcolorbox}
\vspace{-2mm}
\caption{One \lite{} hit and its matched future paper. Generation saw only pre-Nov-2024 papers; the match appeared 15~months later. The hypothesis fused four pre-window papers (H-DPO, SEE-DPO, sharp KL analysis, adversarial uncertainty estimation) that subsequent work converged on. More cases in Appendix~\ref{app:cases}.}
\label{fig:hit-case}
\end{figure}

%% file: sections/04failures.tex
\section{Failure Modes}
\label{sec:failures}

Three boundary conditions surfaced as we used the workflow at scale; we present each as a discovered trade-off, paired with a concrete \emph{detection signal} a user can compute on their own runs to decide when each regime applies. Figure~\ref{fig:density} previews the mechanism behind F1 below.

\paragraph{F1: A quality-coverage trade-off in change-detection instrumentation.}
We tested an instrumented variant (\full{} in Table~\ref{tab:main}) that adds an explicit change-detection step classifying each window-level update into five canonical types and feeds the labels into the generation prompt.
Compared to \lite{}, this variant trades coverage for per-hit quality: hit rate moves from $5.8\%$ to $1.4\%$ at $11.5\times$ the token cost, while originality and cross-field synthesis (Table~\ref{tab:main}, right block) and per-hit alignment (best-match $3.90\!\to\!4.58$; lead time $404\!\to\!537$ days) all improve.
The mechanism is visible in Figure~\ref{fig:density}: explicit change-tagging concentrates hypotheses around a narrower, higher-quality set, while \lite{}'s broader distribution carries more right-tail mass across the hit threshold.
\textbf{Detection signal}: run any new instrumentation in parallel with \lite{} on a held-out subset; revert if topic coverage drops by $\geq 2\times$.

\paragraph{F2: Predictive accuracy varies with topic stability.}
Embedding window-level hypothesis centroids and measuring cosine drift between first and last window, $8$ of the $10$ hit topics fall below the median drift; per-category, the workflow performs strongest on stable areas (multilingual NLP) and is most challenged on fast-shifting LLM applications, where window vocabulary churn (new benchmarks, new model families) keeps the frontier moving faster than the workflow can anticipate.
\textbf{Detection signal}: compute first-to-last centroid drift per topic; topics above our median ($0.27$) warrant additional scrutiny of generated hypotheses.

\paragraph{F3: Convergence-triggered hypotheses are the most predictive.}
\textsc{Convergence}-triggered hypotheses (multiple recent papers reach similar conclusions) hit at $3.8\%$, while \textsc{Contradiction}-triggered ones (new evidence conflicts with prior findings) hit at $0.8\%$ (roughly $5\times$ lower), even though novelty scores are essentially identical across trigger types.
The asymmetry reflects intrinsic predictability of the underlying literature signal: convergence narrows the field's next step, whereas contradictions mark genuine open uncertainty rather than a clear trajectory.
\textbf{Detection signal}: tag each hypothesis with its trigger type at generation; when prioritizing follow-up experiments, sort \textsc{Convergence} and \textsc{Trend\_Confirmed} ahead of \textsc{Contradiction}.

%% file: sections/05conclusion.tex
\section{Conclusion}
\label{sec:conclusion}
We presented \ckm{}, an AI workflow that processes evolving literature through sliding windows and emits ranked, structured hypotheses scored against papers published after the generation window. On a 50-topic ML benchmark, \lite{} attains a $5.8\%$ future-literature hit rate at $\sim$\$3 per topic, $36/50$ topic coverage, and mean temporal lead $\sim$$404$ days. Three regime boundaries (\S\ref{sec:failures}) qualify these results: instrumentation trades coverage for per-hit quality (F1); high-drift topics resist prediction (F2); convergence signals predict where contradiction signals do not (F3).

We read \ckm{} as one early operationalisation of a broader class of AI-assisted research systems~\citep{lu2025aiscientist, gottweis2025towards, Asai2026OpenScholar} built on incremental ingestion, contrastive integration, and continuous refresh; homeostasis, cross-topic adaptation, and propagation into automated experimentation remain open (Appendix~\ref{app:limitations}). More broadly, predictive validation against future literature offers a falsifiable, low-cost alternative to the contemporary-judge protocols dominant in this space, and applies wherever a corpus has dated publication records.

%% file: sections/06impact.tex
\section*{Impact Statement}

This paper presents an AI workflow that helps machine-learning researchers scope research directions from recent literature.
Two impact considerations seem worth flagging beyond the standard statement.

First, literature-based hypothesis generation can shape what researchers choose to work on.
Because \lite{} ranks hypotheses by alignment with where the field is heading, rather than by intrinsic merit, uncritical use could amplify herding around already-popular directions and disadvantage genuinely novel work that has no near-term followers.
The failure modes we report (\S\ref{sec:failures}) are partly meant to discourage this: in particular, F3 shows that the workflow predicts well exactly when the field is converging, which is also when piling on adds the least marginal value.

Second, the workflow consumes external API capacity and depends on closed commercial models.
We have tried to keep per-topic cost low ($\sim$\$2--3) so that the workflow remains accessible to researchers without industrial-scale compute, and we report exact run costs so that practitioners can budget honestly.
Beyond these considerations, this paper presents work whose goal is to advance the field of Machine Learning, and we do not see specific additional societal consequences that warrant highlighting here.

%% file: sections/07appendix.tex
\section{Related Work}
\label{app:related}

\paragraph{LLM-driven scientific discovery.}
The literature has moved through three rough waves over the last two years, all sharing one structural property: validation against a contemporary judge rather than against the future literature itself (Table~\ref{tab:related}).
The first wave emits standalone research \emph{ideas}: SciMON \citep{wang-etal-2024-scimon} mines neighborhood relations between concepts and validates with automatic scoring plus human review; ResearchAgent \citep{baek2025researchagent} adds an agent peer-review loop on top of generated ideas and plans; Nova \citep{hu2025nova} runs a Swiss-style tournament; Scideator \citep{radensky2024scideator} recombines facets of existing papers in human-in-the-loop fashion; Chain-of-Ideas \citep{li2025chain} chains LLM agents through iterative refinement.
The second wave commits to \emph{hypotheses}: AI Co-Scientist \citep{gottweis2025towards} validates with wet-lab experiments (the strongest correctness signal in this set, though limited to a single domain), and MOOSE-Chem2 \citep{yang2026moose} uses hierarchical search against known chemistry targets, both building on the LLM-as-hypothesis-generator paradigm of \citet{zhou-etal-2024-hypothesis} and \citet{yang2024large}, grounded in scientific knowledge by \citet{xiong2024improving}.
The third wave generates \emph{full papers}: AI Scientist \citep{lu2025aiscientist} self-evaluates its outputs; Dolphin \citep{yuan2025dolphin} closes a thinking-practice-feedback loop; CycleResearcher \citep{weng2025cycleresearcher} pairs generation with simulated peer review; EvoScientist \citep{lyu2026evoscientist} adds multi-agent evolution with human peer review.
These systems are more ambitious in scope but inherit the contemporary-judge limitation.

\paragraph{Foundations and adjacent work.}
The argument that anticipation matters in scientific reasoning is not new.
Swanson's classical work on bridging disjoint biomedical literatures \citep{Swanson1986-SWAFOR,swanson1988migraine,Smalheiser1998UsingAA} long ago made the case that connecting evolving findings is itself a form of discovery; modern temporal knowledge-graph methods \citep{cai2024survey,cai2022temporal} carry that idea forward at the level of facts and triples; and knowledge-editing work \citep{decao2021editing,mitchell2022fast} addresses the orthogonal problem of overwriting model facts.
Recent benchmarks IdeaBench \citep{guo2025ideabench} and LiveIdeaBench \citep{ruan2026evaluating} score generated ideas under minimal context, but again with contemporary judges.

\paragraph{How \ckm{} differs.}
\ckm{} fills the gap left by all of this work along two axes.
\emph{On validation}, instead of asking a contemporary judge whether an output looks good, we ask whether subsequent literature actually pursued the predicted direction, turning hypothesis generation into a falsifiable forecasting task.
\emph{On generation}, a sliding-window state $\mathcal{K}_t$ is updated incrementally, so each window's hypotheses condition on what just \emph{changed} in the literature rather than on a static snapshot.
The empirical consequences of these two choices, both where they help and where they hurt, are the body of the paper.

\begin{table}[!htbp]
\centering
\footnotesize
\setlength{\tabcolsep}{6pt}
\begin{tabular}{@{}llll@{}}
\toprule
\textbf{System} & \textbf{Output} & \textbf{Validation} & \textbf{Predictive?} \\
\midrule
SciMON \citep{wang-etal-2024-scimon}        & ideas         & auto + human          & \boldmath$\times$ \\
ResearchAgent \citep{baek2025researchagent} & ideas+plans   & agent peer review     & \boldmath$\times$ \\
Nova \citep{hu2025nova}                     & ideas         & Swiss tournament      & \boldmath$\times$ \\
Scideator \citep{radensky2024scideator}     & ideas         & human-in-the-loop     & \boldmath$\times$ \\
Chain-of-Ideas \citep{li2025chain}          & ideas         & LLM critique          & \boldmath$\times$ \\
AI Co-Scientist \citep{gottweis2025towards} & hypotheses    & wet-lab (biomed)      & \boldmath$\times$ \\
MOOSE-Chem2 \citep{yang2026moose}       & hypotheses    & known-target match    & \boldmath$\times$ \\
AI Scientist \citep{lu2025aiscientist}      & papers        & self-evaluation       & \boldmath$\times$ \\
Dolphin \citep{yuan2025dolphin}             & papers        & benchmark empirical   & \boldmath$\times$ \\
CycleResearcher \citep{weng2025cycleresearcher} & papers    & simulated review      & \boldmath$\times$ \\
EvoScientist \citep{lyu2026evoscientist}    & papers        & peer review + experts & \boldmath$\times$ \\
\midrule
\textbf{\ckm{} (ours)} & \textbf{hypotheses} & \textbf{future-paper alignment} & \boldmath$\checkmark$ \\
\bottomrule
\end{tabular}
\caption{Representative LLM-based scientific-discovery systems and their validation methods. All prior systems validate against contemporary judgments (human reviewers, agent peer review, simulated tournaments, lab assays, benchmark targets, or self-assessment) rather than literature that appears \emph{after} the generation window. \ckm{} validates predictively against future literature.}
\label{tab:related}
\end{table}

\section{Benchmark Topics}
\label{app:topics}

The 50 benchmark topics were chosen to span eight ML categories with deliberately heterogeneous evolution rates, so that the workflow is stress-tested under both stable (mature subfields) and fast-moving (recent LLM topics) conditions.
For every topic the underlying arXiv corpus satisfies three temporal-coverage criteria: $\geq 30$ papers in $[2019, 2024]$ for initialization, $\geq 50$ papers in $[2024, 2025]$ for evolution, and $\geq 30$ papers in $[2025, 2027]$ for validation, with $\geq 100$ papers total per topic.
Topics that fail any criterion are excluded from the sampling pool.

\paragraph{Multilingual \& speech (4).}
Low-resource machine translation; cross-lingual transfer for low-resource languages; multilingual large language models; low-resource speech recognition.

\paragraph{LLM core capabilities (10).}
Chain-of-thought reasoning in LLMs; long-context understanding in LLMs; factual consistency in LLMs; prompt engineering and in-context learning; instruction tuning for LLMs; complex reasoning for AI agents; tool-using agents with LLMs; accuracy of automated evaluation for LLMs; reinforcement learning from human feedback; code generation with LLMs.

\paragraph{LLM applications (8).}
AI for software engineering; security of code LLMs; document understanding and information extraction; aspect-based sentiment analysis; relation extraction from text; open-domain question answering with LLMs; diversity evaluation in text generation; conversational information retrieval.

\paragraph{Safety \& trust (4).}
Explainability and interpretability of neural networks; adversarial robustness of deep learning models; fairness and bias mitigation in machine learning; out-of-distribution detection for language models.

\paragraph{Multimodal (3).}
Vision-language models and multimodal learning; visual question answering with multimodal models; text-to-image generation and diffusion models.

\paragraph{Domain-specific AI (6).}
Clinical natural language processing; machine learning for drug discovery; deep learning for medical image analysis; protein structure prediction with deep learning; surrogate modeling in physics-informed machine learning; AI for hypothesis generation in science.

\paragraph{Efficiency \& data (7).}
Efficient fine-tuning of large language models; knowledge distillation for small language models; model compression and pruning for neural networks; mixture of experts routing for language models; data filtering for domain-specific models; synthetic data quality evaluation; data augmentation for natural language processing.

\paragraph{Other foundations (8).}
Continual learning and catastrophic forgetting; domain adaptation and transfer learning; federated learning for language models; knowledge graph completion and reasoning; graph neural networks for natural language processing; active learning for natural language processing; deep learning for recommendation systems; user experience evaluation for large language models.

\section{Trigger-Type Breakdown (extended F3)}
\label{app:triggers}

F3 in \S\ref{sec:failures} reports a $5\times$ gap between \textsc{Convergence} and \textsc{Contradiction} triggers, but the change-detection step actually classifies each window-level update into a richer set of types. We list them here, with the per-trigger rationale and the full trigger~$\times$~drift breakdown across the $892$ \full{} hypotheses.

\paragraph{Trigger taxonomy.}
\begin{itemize}[leftmargin=4mm,itemsep=1pt,topsep=2pt]
    \item \textsc{Convergence}: multiple recent papers independently reach similar conclusions, signalling a coalescing direction.
    \item \textsc{Trend\_Confirmed}: a previously emerging trend is corroborated by additional within-window evidence.
    \item \textsc{Bridge}: a finding connects two previously disjoint subareas of the knowledge state.
    \item \textsc{Non\_Obvious\_Bridge}: same as \textsc{Bridge} but across distant subareas (different arXiv categories or vocabularies).
    \item \textsc{Gap}: a previously identified open question receives partial new evidence without resolution.
    \item \textsc{Gap\_Exploitation}: an open question is directly addressed by an in-window paper, suggesting the gap is now actively being filled.
    \item \textsc{Contradiction}: a new finding directly conflicts with a claim already in $\mathcal{K}_{t-1}$.
    \item \textsc{Cross\_Paper}: a single new paper synthesizes elements from multiple prior $\mathcal{K}_{t-1}$ subareas without producing a new direction.
    \item \textsc{Incremental}: routine refinement; included for completeness but rarely fires.
\end{itemize}

\paragraph{Trigger $\times$ topic-drift hit rates.}
Splitting topics at the median first-to-last centroid drift ($0.267$, used in F2) and crossing with trigger type gives a non-uniform structure (Table~\ref{tab:trigger_drift}).
The most striking cell is \textsc{Gap\_Exploitation} in low-drift topics ($n{=}9$, $33.3\%$ hit rate); the small sample warrants caution, but the absence of any hits in the corresponding high-drift cell suggests that gap-filling is genuinely a regime-dependent signal rather than a noise artifact.
Low-drift topics account for $10$ of the $12$ total \full{} hits, consistent with F2.

\begin{table}[!htbp]
\centering
\small
\setlength{\tabcolsep}{6pt}
\begin{tabular}{lcccc}
\toprule
& \multicolumn{2}{c}{\textbf{Low Drift}} & \multicolumn{2}{c}{\textbf{High Drift}} \\
\cmidrule(lr){2-3} \cmidrule(lr){4-5}
\textbf{Trigger} & $n$ & Hit\% & $n$ & Hit\% \\
\midrule
\textsc{Gap\_Exploitation} & 9   & \textbf{33.3} & 18  & 0.0 \\
\textsc{Bridge}            & 104 & 2.9           & 77  & 0.0 \\
\textsc{Gap}               & 73  & 2.7           & 72  & 1.4 \\
\textsc{Non\_Obv.\_Bridge} & 69  & 1.4           & 103 & 1.0 \\
\textsc{Contradiction}     & 81  & 1.2           & 75  & 0.0 \\
\textsc{Cross\_Paper}      & 36  & 0.0           & 54  & 0.0 \\
\bottomrule
\end{tabular}
\caption{Hit rate by trigger $\times$ topic drift across all $892$ \full{} hypotheses. Drift split at the topic-level median ($0.267$). \textsc{Gap\_Exploitation} in low-drift topics achieves the highest hit rate, although the sample is small. \textsc{Cross\_Paper} and any combination with high drift reach $0\%$, suggesting that the \emph{interaction} of trigger type with topic stability is what governs predictability, not either alone.}
\label{tab:trigger_drift}
\end{table}

\paragraph{Practical takeaway.}
A practitioner deploying \lite{} can use this table as a triage rule: among hypotheses generated in a topic with low semantic drift across windows, prioritize \textsc{Gap\_Exploitation} and \textsc{Bridge} triggers; skip \textsc{Cross\_Paper} triggers and any high-drift conditions for time-constrained reading. This is the operational form of F3, generalized beyond the headline convergence-vs-contradiction split in the main body.

\section{Hypothesis Trajectories Across Windows}
\label{app:trajectories}

\begin{figure}[!t]
    \centering
    \includegraphics[width=0.78\linewidth]{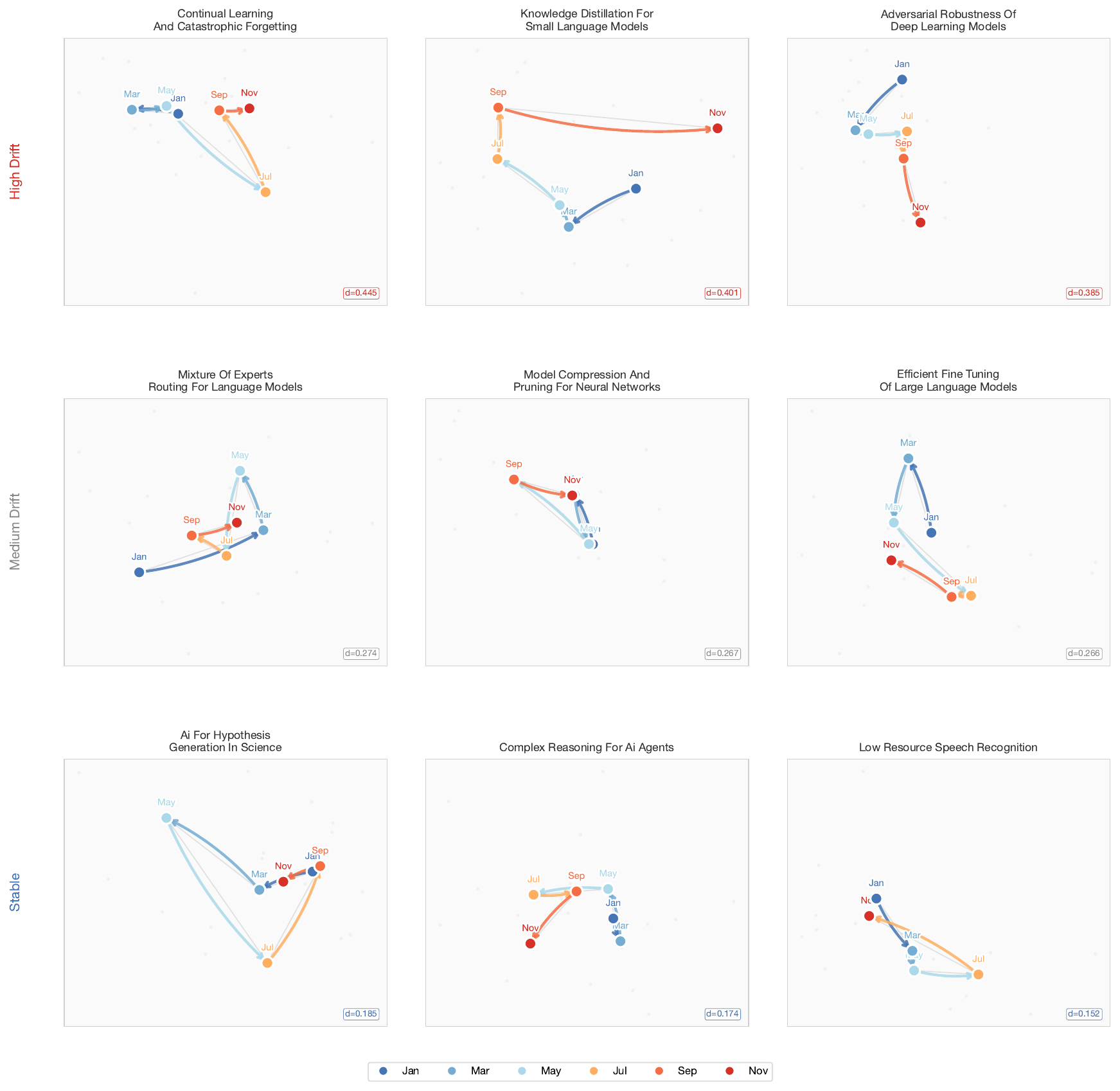}
    \caption{Hypothesis-centroid trajectories across six evolution windows for nine representative topics, PCA-projected from sentence embeddings. Each colored marker is a window's hypothesis-set centroid; the polyline is the temporal sequence. Topics in the top row are low-drift (and produce predictive hits); topics in the bottom row are high-drift (and tend not to). This is the visual evidence behind F2 in \S\ref{sec:failures}: $8$ of the $10$ hit topics fall below the median drift.}
    \label{fig:trajectory}
\end{figure}

Figure~\ref{fig:trajectory} visualizes the literal motion of \ckm{}'s hypothesis centroids across the six 2-month windows for nine representative topics, in PCA-projected embedding space. This is the empirical ground truth behind F2 in \S\ref{sec:failures}: low-drift trajectories (top row) tend to produce hits, while high-drift trajectories (bottom row) tend not to. It is also the visual instantiation of the metabolism metaphor introduced in \S\ref{sec:method}: the workflow does not deliver one batch of outputs but produces a moving target whose shape across time is itself diagnostic.

\section{Limitations and Future Work}
\label{app:limitations}

\paragraph{Hit rate vs.\ coverage: which metric matters.}
A $5.8\%$ hit rate at $17$ hypotheses per topic produces, on average, roughly one validated direction per topic, which in the deployment scenario this work targets (a researcher scoping 4--5 candidate subareas in parallel before committing to one) is precisely the useful unit of output. The substantive question is therefore not what fraction of generated hypotheses hit, but whether the researcher reliably gets at least one valid prediction per topic of interest. Coverage ($36/50 = 72\%$ for \lite{}, against $15/50 = 30\%$ for the one-shot baseline) maps directly to that question; the per-hypothesis hit rate is its dual and is mechanically diluted by hypothesis yield, so a low headline rate at high yield does not by itself indicate weak prediction. We recommend that future evaluations of literature-based hypothesis generation report both metrics, and that comparisons of competing systems privilege coverage when deployment to a portfolio of topics is the use case.

\paragraph{Temporal lead distribution.}
Across the $55$ validated hits over $36$ topics, validated hypotheses precede their matched papers by a mean of $404$ days and a median of $399$ days (stdev $174$, range $66$--$757$).
Aggregated to the topic level, the mean per-topic average lead is $411$ days (median $424$, stdev $160$).
The headline RLHF case ($460$ days) sits within this distribution rather than being a tail, supporting the framing that predictive lead is a property of the workflow rather than of a single example.

\paragraph{The validation proxy.}
We treat ``a future paper pursued the same direction'' as evidence that a hypothesis was good.
This rewards predictions that align with where the field went and penalizes genuinely novel directions the field has not yet visited.
Hits should be read as a lower-bound measure of usefulness, not an upper bound on novelty.

\paragraph{Domain scope.}
All 50 topics are within ML.
We have not tested whether the workflow transfers to mathematics, theory, or wet-lab biology, where evidence units, citation patterns, and time scales differ; we leave cross-domain transfer to future work and do not extrapolate from these ML results.

\paragraph{Absence of human evaluation.}
We measure usefulness via future-paper alignment, not via researcher judgement of whether a generated hypothesis would change a project plan.
A complementary user study (e.g., asking $\sim10$ ML researchers to score generated hypotheses for novelty, feasibility, and relevance to their current work) would directly probe practical utility; it is the most natural complement to the predictive evaluation we report and the next study we plan to run.

\paragraph{The workflow does not replace expert judgement.}
Every claim in the main body is about ranked output quality on a benchmark.
Treating any individual hypothesis as scientifically valid still requires a domain expert reading the cited papers and the proposed experiment.

\paragraph{Future work.}
Three directions follow naturally.
\emph{(i)~Domain transfer.} Repeat the protocol on math, theory, and biomedicine to test which failure modes generalize and which are ML-specific; a public benchmark on this axis would be a community resource in itself.
\emph{(ii)~Scale and statistical strength.} Move from $50$ topics to several hundred, which would let F2 (drift sensitivity) be tested at $p < 0.01$ rather than as a qualitative pattern, and would expose lower-frequency trigger types that we currently lack power to characterize.
\emph{(iii)~From metabolism to homeostasis.} \ckm{} operationalizes one living-system property. The next layer is homeostasis under contradicting evidence (revising prior hypotheses when new conflicting findings arrive) and propagation (feeding promising hypotheses into automated experiment runners). Together these would close the loop from literature monitoring to autonomously-tested predictions.

\section{Reproducibility}
\label{app:repro}

Every number in this paper can be regenerated end-to-end on a laptop with API access, in roughly two days of wallclock and at a cost dominated by the API bill rather than the engineering effort.
\lite{} ships as an installable plugin (\url{https://github.com/TaoJinkai/ckm-hypogen}), and a complete single-topic run takes three commands: installing the plugin, launching its host runtime, and invoking the workflow's entry-point command with a topic name (e.g.\ ``long-context modeling'').
Defaults: 2-month windows, up to 96 papers per window, Gemini-2.5-Flash for generation, GPT-4o for judging.
A single-topic run completes in $\sim\!50$ minutes wallclock and $\sim\!\$2$--$3$ in API cost; the full 50-topic benchmark required $44.4$ hours and $\sim\!\$130$.
Exact prompt templates, the topic list, the per-topic run logs that produced every number above, the judging code, and a one-command verification script are in the code repository (\url{https://github.com/TaoJinkai/ckm-hypogen}).

\paragraph{Estimated time savings vs.\ the manual baseline.}
A graduate student doing frontier scoping by hand on one ML subarea (reading 30--60 papers, maintaining cross-session notes, drafting candidate research questions concrete enough to act on) takes 12--20 hours; \lite{} produces a comparable ranked list of 17 structured hypotheses in $\sim$50 minutes wallclock at \$2--3 in API spend, a $15\times$--$25\times$ time reduction per topic. The cognitive task that actually motivates this work is doing this for 4--5 candidate directions \emph{in parallel} before committing a quarter to one of them, where the manual baseline runs out of working hours; under continuous-deployment mode (above), \lite{} stays within \$10--15/day for a researcher tracking that many topics in the background.

\paragraph{Installation and minimal API call.}
The released repository ships with (i) a Python package installable via \texttt{pip install -e benchmark/} (Python $\geq$3.9; dependencies: \texttt{openai}, \texttt{requests}, \texttt{scipy}, \texttt{numpy}, \texttt{python-dotenv}), (ii) a \texttt{.env.example} listing required keys (\texttt{OPENAI\_API\_KEY} for the judge, plus the generator-side key), and (iii) a \texttt{python -m ckm\_benchmark.recompute --summary results/lite\_summary.json} command that reproduces the Table~\ref{tab:main} numbers in seconds without any API calls. A minimal API-driven run on one topic uses the same entry-point command as the full benchmark, with the topic name supplied on the command line; the runner, prompt files, and per-stage logging are all contained in the repository under permissive license.

\paragraph{Topic list.}
50 topics across NLP core (9), LLM methods (10), LLM applications (10), domain-specific AI (6), safety \& ethics (5), multilingual NLP (3), multimodal (2), and other (5).

\paragraph{Per-topic timing.}
Median wallclock $47.5$ min; mean generation tokens $0.51$M.
Per-topic timing variance is dominated by paper full-text resolution latency, not generation.

\paragraph{Continuous deployment for daily research tracking.}
The 2-month window width used in the benchmark is an evaluation choice that lets us complete six windows over the 2024--2025 evolution period within a single laptop-driven run; it is not a deployment requirement.
In production, a user runs \lite{} as a daily or weekly background process on a single research topic, ingesting whatever papers appeared since the last run and emitting new hypotheses against an ever-growing knowledge state.
At a typical arrival rate of one or two ML papers per day per subarea, the per-day token budget drops below \$0.10, putting literature-driven hypothesis tracking within reach of a graduate student following two or three parallel subareas (the bottleneck \S\ref{sec:intro} identified).
Under this deployment mode the metabolism metaphor of \S\ref{sec:method} translates literally: the workflow becomes a researcher's continuously running daily reading routine, and the knowledge state plus hypothesis log accumulates over months as a personal scientific notebook.

\paragraph{Model choices and leakage prevention.}
Generation uses Gemini-2.5-Flash \citep{comanici2025gemini25pushingfrontier}; two-stage judging uses GPT-4o-mini at threshold $5.0$ followed by GPT-4o at threshold $6.0$ \citep{openai2024gpt4ocard}.
All three models have training cutoffs in 2024 or earlier (the GPT-4o family has an October 2023 cutoff, and Gemini-2.5-Flash predates Q2 2025), so the bulk of the 2025--2027 validation literature post-dates their training corpora; this is the temporal half of leakage prevention.
The cross-provider half (Google for generation, OpenAI for judging, so the two never share weights) and the prompt-level safeguard (generation sees only pre-window literature) are documented in Appendix~\ref{app:safeguards}.
Together these rule out the standard memorization paths: the generator cannot have seen the matched future papers at training time, the judge cannot retrieve them through self-preference, and the prompt itself never contains them.

We deliberately do not use the most capable models available at the time of writing (e.g., GPT-5, Claude Opus 4.x, Gemini 3.x).
Two reasons.
First, the Flash + 4o pair keeps per-topic cost in the \$2--3 range and the full benchmark under \$130, which matches the workshop's academic-budget envelope; substituting frontier models would push the per-topic spend an order of magnitude higher without proportional gain on a workflow whose bottleneck is structure rather than raw generation quality.
Second, and more importantly, the workflow is intentionally model-agnostic: its structural choices (sliding-window state, contrastive update, structured hypothesis output, embedding-similarity validation) are independent of any particular generator, and any of the three roles (extraction, generation, judging) can be swapped to a stronger LLM without changing the workflow.
The numbers in this paper should therefore be read as a \emph{lower bound} on what the same workflow achieves under future deployments using frontier-grade models.

\paragraph{Reproducing every Table~\ref{tab:main} number.}
Each per-topic record in the released JSON summaries (one for \lite{}, one for \full{}, one for \batch{}) contains \texttt{duration}, \texttt{total\_generation\_tokens}, \texttt{yield}, \texttt{hit\_rate}, and \texttt{best\_match\_score}. The released code repository includes a single Python script that regenerates every value in Table~\ref{tab:main} as well as the Wilcoxon statistic from \S\ref{sec:eval}, by reading those JSONs and printing per-condition rows.

\paragraph{Wilcoxon paired test details.}
The hit-rate comparison between \lite{} and \batch{} is paired by topic ($N=50$).
After dropping $12$ ties, the test runs on $n_{\mathrm{eff}}=38$ non-zero differences with $W=182.5$, normal-approximation $|Z|=2.73$, and $p=0.006$ (two-sided).
The matched-pair effect size $r=|Z|/\sqrt{n_{\mathrm{eff}}}=0.44$ is in the medium-to-large range; \lite{} beats \batch{} on $29$ topics, loses on $9$, and ties on $12$.
The full computation is reproduced by the same one-command verification script.

\section{Validity Safeguards}
\label{app:safeguards}
(1) Generation prompts see only pre-window literature.
(2) Validation papers are drawn from 2025--2027, post-training for the generation model.
(3) Generation and judging use different model families.
(4) Two-stage judging with the stronger judge at the higher threshold.
(5) Embedding pre-filter is model-agnostic and only widens the candidate pool.

\section{Per-hypothesis Novelty Judge}
\label{app:novelty}

The novelty scores in the right block of Table~\ref{tab:main} are produced by a separate, single-stage LLM judge (GPT-4o, distinct from the hit-verification judge of Appendix~\ref{app:safeguards}) that scores each generated hypothesis on four standalone $1$--$10$ dimensions. The judge sees only the hypothesis text (statement, research claim, source-paper citations) and never the validation pool, so novelty scoring is independent of predictive hit assessment. The four dimensions:

\begin{itemize}[leftmargin=4mm,itemsep=2pt,topsep=2pt]
    \item \textbf{D1 Originality.} How original is the proposed method delta relative to the cited source papers? Low scores (1--3) mark hypotheses that essentially restate one cited paper; high scores (8--10) mark hypotheses that propose a non-obvious recombination or extension that no cited source authored alone.
    \item \textbf{D2 Cross-field synthesis.} Does the hypothesis bridge two or more subareas of the knowledge state, or distinct arXiv categories? Low scores mark single-subarea hypotheses; high scores mark genuinely cross-subarea proposals (e.g., applying a technique from one ML community to a problem in another).
    \item \textbf{D3 Gap precision.} How precisely does the hypothesis identify the literature gap it claims to fill? Low scores mark vague gap statements (``X is an open problem''); high scores mark gaps localised to a specific failure mode of a specific prior method, with that method named.
    \item \textbf{D4 Falsifiability.} Does the hypothesis specify an experiment that could refute it? Low scores mark unfalsifiable claims; high scores mark hypotheses that name a target setting, baseline, expected observable, and evaluation plan concretely enough that an experiment can be staged.
\end{itemize}

D4 is uniformly high across systems ($7.7$--$7.9$) because the structured-output prompt requires every hypothesis to populate the seven-field research-claim schema (Appendix~\ref{app:worked-example}); falsifiability is therefore driven by the prompt template, not by the generation strategy. D1, D2, D3 differ across systems and are what Table~\ref{tab:main} (right block) actually contrasts.

\section{Prompt Templates}
\label{app:prompts}
We summarize the structural commitments of the three workflow stages (initialization, window update, hypothesis generation) and the two judging stages (GPT-4o-mini pre-filter, GPT-4o re-judgment) below.

\paragraph{Stage 1 (Initialization) prompt.} The LLM is given a topic query, a list of $\leq 48$ pre-window arXiv papers (title, abstract, full text), and instructions to extract per-paper structured notes (core method, key findings, datasets, open questions) and to organize them into Markdown topic files partitioned by sub-theme.

\paragraph{Stage 2 (Window update) prompt.} The LLM is given the prior topic files (the current $\mathcal{K}_{t-1}$), the in-window papers $\mathcal{P}_t$, and instructions to produce the updated $\mathcal{K}_t$: append novel claims, mark confirming evidence, and flag contradictions where new findings conflict with $\mathcal{K}_{t-1}$.

\paragraph{Stage 3 (Hypothesis generation) prompt.} The LLM is given $\mathcal{K}_t$, a running summary of cross-window changes $\mathcal{S}_{1:t-1}$, the cumulative prior hypotheses $\mathcal{H}_{1:t-1}$, and instructions to emit a small batch of new hypotheses in the structured JSON schema documented in Appendix~\ref{app:worked-example}.

\paragraph{Judging prompts.} Both judges receive a candidate hypothesis and a candidate future paper (title + abstract + first 2000 characters of body) and return a $1$--$10$ alignment score with a one-paragraph reasoning. The pre-filter (GPT-4o-mini) accepts candidates above $5.0$; the re-judge (GPT-4o) confirms hits above $6.0$.

\section{Additional Hit Examples}
\label{app:cases}

The main paper highlights one hit (Figure~\ref{fig:hit-case}, RLHF stability). We present three additional cases below, sampled to span different research communities and trigger types. In each box the workflow saw only papers published before the listed generation date, and the matched paper appeared after.

\begin{tcolorbox}[
  colback=white, colframe=black!35, boxrule=0.5pt, arc=1.5mm,
  left=5pt, right=5pt, top=4pt, bottom=4pt,
  title={\small\bfseries Case A: OOD Detection for LLMs \hfill Trigger: \textsc{Convergence} $\cdot$ Generated May 2024 $\cdot$ \textcolor{green!50!black}{Hit $\checkmark$ \textbf{280-day lead}}},
  fonttitle=\small, coltitle=black, colbacktitle=gray!10,
]
\begin{tcolorbox}[
  colback=teal!4, colframe=teal!50, boxrule=0.3pt, arc=1mm,
  left=6pt, right=6pt, top=4pt, bottom=4pt,
]
\small\itshape
``Leveraging large language models to generate diverse textual outlier samples for a given in-distribution text classification task, by prompting the LLM to create semantically distinct but plausible OOD examples, will significantly improve OOD detection performance (AUROC, FPR@95) compared to methods relying on synthetic outliers derived solely from ID data, particularly for near-OOD scenarios.''
\end{tcolorbox}
\vspace{2pt}
\small
\textbf{Matched (Feb 2025)} $\rightarrow$ \emph{Out-of-Distribution Detection using Synthetic Data Generation} (arXiv:2502.03323), which independently proposed an LLM-driven synthetic-OOD pipeline targeting the same near-OOD AUROC/FPR@95 metrics named in the hypothesis.
\end{tcolorbox}

\vspace{6pt}

\begin{tcolorbox}[
  colback=white, colframe=black!35, boxrule=0.5pt, arc=1.5mm,
  left=5pt, right=5pt, top=4pt, bottom=4pt,
  title={\small\bfseries Case B: Long-context Multimodal Modeling \hfill Trigger: \textsc{Convergence} $\cdot$ Generated March 2024 $\cdot$ \textcolor{green!50!black}{Hit $\checkmark$ \textbf{326-day lead}}},
  fonttitle=\small, coltitle=black, colbacktitle=gray!10,
]
\begin{tcolorbox}[
  colback=teal!4, colframe=teal!50, boxrule=0.3pt, arc=1mm,
  left=6pt, right=6pt, top=4pt, bottom=4pt,
]
\small\itshape
``A multimodal LLM architecture that combines hierarchical token merging for local segment features with a dynamic, compressed memory bank for long-term historical context will achieve superior fine-grained understanding and temporal consistency in long-term video question answering and captioning, while maintaining efficient GPU memory usage.''
\end{tcolorbox}
\vspace{2pt}
\small
\textbf{Matched (Jan 2025)} $\rightarrow$ \emph{InternVideo2.5: Empowering Video MLLMs with Long and Rich Context Modeling} (arXiv:2501.12386), which uses hierarchical token compression with a memory mechanism for long-video understanding; both structural choices named in the hypothesis.
\end{tcolorbox}

\vspace{6pt}

\begin{tcolorbox}[
  colback=white, colframe=black!35, boxrule=0.5pt, arc=1.5mm,
  left=5pt, right=5pt, top=4pt, bottom=4pt,
  title={\small\bfseries Case C: Knowledge Graph Completion \hfill Trigger: \textsc{Convergence} $\cdot$ Generated September 2024 $\cdot$ \textcolor{green!50!black}{Hit $\checkmark$ \textbf{$\sim$120-day lead}}},
  fonttitle=\small, coltitle=black, colbacktitle=gray!10,
]
\begin{tcolorbox}[
  colback=teal!4, colframe=teal!50, boxrule=0.3pt, arc=1mm,
  left=6pt, right=6pt, top=4pt, bottom=4pt,
]
\small\itshape
``A novel inductive KGC framework that integrates LLM-based reasoning with explicit Context Graphs by leveraging latent type constraints and degree-based filtered reasoning paths.''
\end{tcolorbox}
\vspace{2pt}
\small
\textbf{Matched (Jan 2025)} $\rightarrow$ \emph{KG-CF: Knowledge Graph Completion with Context Filtering under the Guidance of LLMs} (arXiv:2501.02711), which combines LLM reasoning with context-graph filtering using type and degree constraints; all three structural elements named in the hypothesis.
\end{tcolorbox}

\paragraph{What these cases share.}
Across the four hits documented in this paper (RLHF in the main body and Cases A--C here), three patterns recur. First, all four were triggered by \textsc{Convergence} signals, consistent with F3. Second, the hypothesis text in each case fuses 3--4 pre-window source papers into a single integrated proposal, and the matched future paper realizes essentially that integration: the workflow's value comes less from generating wholly new ideas than from naming the integration the field is about to make. Third, the leads range from $\sim$120 to $\sim$680 days, comfortably above the per-topic median (193 days for the long-context topic alone), indicating that the workflow is not merely guessing the next month but anticipating directions on a multi-quarter horizon.

\section{Using a CKM Hypothesis: A Worked Example}
\label{app:worked-example}

A typical run returns $\sim$17 ranked hypotheses per topic. Each hypothesis is a structured artifact, not a free-form paragraph, and is meant to be \emph{consumed} not merely read. This appendix shows the full schema and walks through how a researcher uses one.

\paragraph{Hypothesis schema.}
Every generated hypothesis follows the structure below. The example fields are filled with the actual content of the RLHF hit highlighted in Figure~\ref{fig:hit-case} (hypothesis ID \texttt{hyp-2024-11-015}, generated November~2024).

\begin{small}
\begin{itemize}[leftmargin=4mm,itemsep=2pt,topsep=2pt]
    \item \textbf{Statement} (one sentence): \emph{``A novel RLHF framework will integrate adaptive entropy regularization (H-DPO/SEE-DPO style) with theoretically grounded KL-regularization and dynamic uncertainty-aware policy optimization, demonstrating enhanced stability and reduced reward hacking compared to standard DPO or PPO-based RLHF.''}

    \item \textbf{Research Claim} (six fields):
    \begin{itemize}[leftmargin=4mm,itemsep=1pt,topsep=2pt]
        \item \textsc{Problem}: RLHF and DPO methods often struggle with reward overoptimization, mode collapse, and instability, particularly in complex generative tasks.
        \item \textsc{Method Delta}: Integrate adaptive entropy regularization with theoretically grounded KL-regularization and dynamic uncertainty-aware policy optimization.
        \item \textsc{Target Setting}: LLM alignment (math, code, instruction following) and text-to-image diffusion alignment.
        \item \textsc{Baseline}: Standard DPO or PPO-based RLHF with fixed KL regularization.
        \item \textsc{Expected Observable}: Smoother reward curves, lower rates of repetitive/nonsensical high-reward outputs, higher pass@k for math/code, better diversity metrics.
        \item \textsc{Evaluation Plan}: Implement and evaluate on GSM8K, HumanEval, MMLU-Pro, IFEval (LLMs) and Pick-a-Pic-V1 (diffusion); ablate the adaptive regularization components.
        \item \textsc{Failure Mode}: Interplay between multiple regularizers may introduce hyperparameter tuning challenges.
    \end{itemize}

    \item \textbf{Source Papers} (the workflow's evidence anchors, all pre-window):
    \begin{itemize}[leftmargin=4mm,itemsep=1pt,topsep=2pt]
        \item arXiv:2411.07595 -- \emph{Entropy Controllable Direct Preference Optimization} (H-DPO).
        \item arXiv:2411.04712 -- \emph{SEE-DPO: Self Entropy Enhanced Direct Preference Optimization}.
        \item arXiv:2411.04625 -- \emph{Sharp Analysis for KL-Regularized Contextual Bandits and RLHF}.
        \item arXiv:2403.05171 -- \emph{Overcoming Reward Overoptimization via Adversarial Policy Optimization with Lightweight Uncertainty Estimation}.
    \end{itemize}

    \item \textbf{Trigger}: \textsc{Convergence}; multiple recent papers independently address RLHF stability via adaptive regularization.

    \item \textbf{Self-Assessment} (1--5): novelty 4, feasibility 3, impact 5.
\end{itemize}
\end{small}

\paragraph{How a researcher consumes one hypothesis.}
We recommend the following five-step protocol, which takes roughly three to five minutes per hypothesis and pairs naturally with the F3 detection signal in \S\ref{sec:failures}:

\begin{enumerate}[leftmargin=5mm,itemsep=1pt,topsep=2pt]
    \item \textbf{Triage by self-assessment} -- drop the hypothesis if feasibility~$<3$ (removes about a third of a typical run).
    \item \textbf{Triage by trigger} -- if the trigger type is \textsc{Contradiction} or \textsc{Cross\_Paper}, deprioritize unless explicitly hunting high-risk directions (per F3 and Appendix~\ref{app:triggers}).
    \item \textbf{Verify via source papers, not via the hypothesis text} -- skim the four cited arXiv abstracts and decide whether the cited papers individually justify the proposed integration. The hypothesis's value lies in naming the integration, not in citing well.
    \item \textbf{Sanity-check the experiment plan} -- the \textsc{Target Setting}, \textsc{Baseline}, and \textsc{Expected Observable} fields should compose into a one-day pilot you would actually run with current infrastructure.
    \item \textbf{Promote or discard} -- two or three hypotheses per topic typically survive this filter; queue them for a 1--2 day pilot and discard the rest.
\end{enumerate}

%% file: main.bbl
\begin{thebibliography}{27}
\providecommand{\natexlab}[1]{#1}
\providecommand{\url}[1]{\texttt{#1}}
\expandafter\ifx\csname urlstyle\endcsname\relax
  \providecommand{\doi}[1]{doi: #1}\else
  \providecommand{\doi}{doi: \begingroup \urlstyle{rm}\Url}\fi

\bibitem[Akari~Asai \& Hajishirzi(2026)Akari~Asai and
  Hajishirzi]{Asai2026OpenScholar}
Akari~Asai, Jacqueline~He, R. S. W. S. A. S. J. C. C. K. L. L. S. S. F. M. D.
  D. W. M. L. J. S. J. D. H. V. K. M. T. P. J. S. L. H. T. B. W. Y. X. L. Z. G.
  N. D. S. W. D. D. W.-t. Y. P. W.~K. and Hajishirzi, H.
\newblock Synthesizing scientific literature with retrieval-augmented language
  models.
\newblock \emph{Nature}, 650:\penalty0 857--863, February 2026.
\newblock \doi{10.1038/s41586-025-10072-4}.
\newblock URL \url{https://doi.org/10.1038/s41586-025-10072-4}.

\bibitem[Baek et~al.(2025)Baek, Jauhar, Cucerzan, and
  Hwang]{baek2025researchagent}
Baek, J., Jauhar, S.~K., Cucerzan, S., and Hwang, S.~J.
\newblock Researchagent: Iterative research idea generation over scientific
  literature with large language models.
\newblock In \emph{Proceedings of the 2025 Conference of the Nations of the
  Americas Chapter of the Association for Computational Linguistics: Human
  Language Technologies (Volume 1: Long Papers)}, pp.\  6709--6738, 2025.

\bibitem[Cai et~al.(2022)Cai, Xiang, Gao, Zhang, Li, and Li]{cai2022temporal}
Cai, B., Xiang, Y., Gao, L., Zhang, H., Li, Y., and Li, J.
\newblock Temporal knowledge graph completion: A survey.
\newblock \emph{arXiv preprint arXiv:2201.08236}, 2022.

\bibitem[Cai et~al.(2024)Cai, Mao, Zhou, Long, Wu, and Lan]{cai2024survey}
Cai, L., Mao, X., Zhou, Y., Long, Z., Wu, C., and Lan, M.
\newblock A survey on temporal knowledge graph: Representation learning and
  applications.
\newblock \emph{arXiv preprint arXiv:2403.04782}, 2024.

\bibitem[De~Cao et~al.(2021)De~Cao, Aziz, and Titov]{decao2021editing}
De~Cao, N., Aziz, W., and Titov, I.
\newblock Editing factual knowledge in language models.
\newblock In \emph{Proceedings of the 2021 Conference on Empirical Methods in
  Natural Language Processing}, pp.\  6491--6506, 2021.

\bibitem[Gemini~Team(2025)]{comanici2025gemini25pushingfrontier}
Gemini~Team, G.
\newblock {Gemini 2.5}: Pushing the frontier with advanced reasoning,
  multimodality, long context, and next generation agentic capabilities, 2025.
\newblock URL \url{https://arxiv.org/abs/2503.21218}.

\bibitem[Gottweis et~al.(2026)Gottweis, Weng, Daryin, Tu, Sirkovic, Myaskovsky,
  Glowaty, Weissenberger, Orlandi, Popovici, Palepu, Rong, Tanno, Saab, Zhang,
  Blum, Carroll, Kulkarni, Tomašev, Zverinski, Rendulic, Vedadi, Hasler,
  Luka~Rimanic, Budiselic, Feinstein, Bellaiche, Sheffer, Freyberg, Ratcliff,
  Bertolli, Chou, Hassidim, Gokturk, Vahdat, Guan, Dhillon, Vaishnav, Lee,
  Costa, Penadés, Peltz, Matias, Manyika, Hassabis, Xu, Kohli, Pawlosky,
  Karthikesalingam, and Natarajan]{gottweis2025towards}
Gottweis, J., Weng, W.-H., Daryin, A., Tu, T., Sirkovic, P., Myaskovsky, A.,
  Glowaty, G., Weissenberger, F., Orlandi, A., Popovici, D., Palepu, A., Rong,
  K., Tanno, R., Saab, K., Zhang, F., Blum, J., Carroll, A., Kulkarni, K.,
  Tomašev, N., Zverinski, D., Rendulic, I., Vedadi, E., Hasler, F.,
  Luka~Rimanic, M.~B., Budiselic, I., Feinstein, B., Bellaiche, M., Sheffer,
  T., Freyberg, J., Ratcliff, J., Bertolli, O., Chou, K., Hassidim, A.,
  Gokturk, B., Vahdat, A., Guan, Y., Dhillon, V., Vaishnav, E.~D., Lee, B.,
  Costa, T. R.~D., Penadés, J.~R., Peltz, G., Matias, Y., Manyika, J.,
  Hassabis, D., Xu, Y., Kohli, P., Pawlosky, A., Karthikesalingam, A., and
  Natarajan, V.
\newblock Accelerating scientific discovery with co-scientist.
\newblock \emph{Nature}, pp.\  1--3, 2026.

\bibitem[Guo et~al.(2025)Guo, Shariatmadari, Xiong, Huang, Kim, Williams,
  Bekiranov, and Zhang]{guo2025ideabench}
Guo, S., Shariatmadari, A.~H., Xiong, G., Huang, A., Kim, M., Williams, C.~M.,
  Bekiranov, S., and Zhang, A.
\newblock Ideabench: Benchmarking large language models for research idea
  generation.
\newblock In \emph{Proceedings of the 31st ACM SIGKDD Conference on Knowledge
  Discovery and Data Mining V. 2}, pp.\  5888--5899, 2025.

\bibitem[Hu et~al.(2025)Hu, Fu, Wang, Wang, Li, Xu, Lu, Jin, Pan, and
  Lan]{hu2025nova}
Hu, X., Fu, H., Wang, J., Wang, Y., Li, Z., Xu, R., Lu, Y., Jin, Y., Pan, L.,
  and Lan, Z.
\newblock Nova: An iterative planning framework for enhancing scientific
  innovation with large language models.
\newblock In \emph{Findings of the Association for Computational Linguistics:
  ACL 2025}, pp.\  21330--21359, 2025.

\bibitem[Li et~al.(2025)Li, Xu, Guo, Zhao, Li, Yuan, Zhang, Jiang, Xin, Dang,
  Rong, Zhao, Feng, and Bing]{li2025chain}
Li, L., Xu, W., Guo, J., Zhao, R., Li, X., Yuan, Y., Zhang, B., Jiang, Y., Xin,
  Y., Dang, R., Rong, Y., Zhao, D., Feng, T., and Bing, L.
\newblock Chain of ideas: Revolutionizing research via novel idea development
  with {LLM} agents.
\newblock In Christodoulopoulos, C., Chakraborty, T., Rose, C., and Peng, V.
  (eds.), \emph{Findings of the Association for Computational Linguistics:
  EMNLP 2025}, pp.\  8971--9004, Suzhou, China, November 2025. Association for
  Computational Linguistics.
\newblock ISBN 979-8-89176-335-7.
\newblock \doi{10.18653/v1/2025.findings-emnlp.477}.
\newblock URL \url{https://aclanthology.org/2025.findings-emnlp.477/}.

\bibitem[Lu et~al.(2026)Lu, Lu, Lange, Yamada, Hu, Foerster, Ha, and
  Clune]{lu2025aiscientist}
Lu, C., Lu, C., Lange, R.~T., Yamada, Y., Hu, S., Foerster, J., Ha, D., and
  Clune, J.
\newblock Towards end-to-end automation of ai research.
\newblock \emph{Nature}, 651\penalty0 (8107):\penalty0 914--919, 2026.

\bibitem[Lyu et~al.(2026)Lyu, Zhang, Yi, Zhao, Shuyu~Guo, Piotrowski, Kaliski,
  Urbani, Meng, Zhou, and Yan]{lyu2026evoscientist}
Lyu, Y., Zhang, X., Yi, X., Zhao, Y., Shuyu~Guo, W.~H., Piotrowski, J.,
  Kaliski, J., Urbani, J., Meng, Z., Zhou, L., and Yan, X.
\newblock Evoscientist: Towards multi-agent evolving ai scientists for
  end-to-end scientific discovery.
\newblock \emph{arXiv preprint arXiv:2603.08127}, 2026.

\bibitem[Mitchell et~al.(2022)Mitchell, Lin, Bosselut, Finn, and
  Manning]{mitchell2022fast}
Mitchell, E., Lin, C., Bosselut, A., Finn, C., and Manning, C.~D.
\newblock Fast model editing at scale.
\newblock In \emph{International Conference on Learning Representations}, 2022.

\bibitem[{OpenAI}(2024)]{openai2024gpt4ocard}
{OpenAI}.
\newblock {GPT-4o System Card}, 2024.
\newblock URL \url{https://arxiv.org/abs/2410.21276}.

\bibitem[Radensky et~al.(2024)Radensky, Shahid, Fok, Siangliulue, Hope, and
  Weld]{radensky2024scideator}
Radensky, M., Shahid, S., Fok, R., Siangliulue, P., Hope, T., and Weld, D.~S.
\newblock Scideator: Human-llm scientific idea generation grounded in
  research-paper facet recombination.
\newblock \emph{arXiv preprint arXiv:2409.14634}, 2024.

\bibitem[Ruan et~al.(2026)Ruan, Wang, Hong, Sun, Wang, and
  Liu]{ruan2026evaluating}
Ruan, K., Wang, X., Hong, J., Sun, H., Wang, P., and Liu, Y.
\newblock Evaluating {LLMs}' divergent thinking capabilities for scientific
  idea generation with minimal context.
\newblock \emph{Nature Communications}, mar 2026.
\newblock \doi{10.1038/s41467-026-70245-1}.
\newblock URL \url{https://doi.org/10.1038/s41467-026-70245-1}.

\bibitem[Si et~al.(2025)Si, Yang, and Hashimoto]{si2025can}
Si, C., Yang, D., and Hashimoto, T.
\newblock Can {LLM}s generate novel research ideas? a large-scale human study
  with 100+ {NLP} researchers.
\newblock In \emph{The Thirteenth International Conference on Learning
  Representations}, 2025.
\newblock URL \url{https://openreview.net/forum?id=M23dTGWCZy}.

\bibitem[Smalheiser \& Swanson(1998)Smalheiser and
  Swanson]{Smalheiser1998UsingAA}
Smalheiser, N.~R. and Swanson, D.~R.
\newblock Using arrowsmith: a computer-assisted approach to formulating and
  assessing scientific hypotheses.
\newblock \emph{Computer methods and programs in biomedicine}, 57\penalty0
  (3):\penalty0 149--153, 1998.

\bibitem[Swanson(1986)]{Swanson1986-SWAFOR}
Swanson, D.~R.
\newblock Fish oil, raynaud's syndrome, and undiscovered public knowledge.
\newblock \emph{Perspectives in Biology and Medicine}, 30\penalty0
  (1):\penalty0 7--18, 1986.
\newblock \doi{10.1353/pbm.1986.0087}.

\bibitem[Swanson(1988)]{swanson1988migraine}
Swanson, D.~R.
\newblock Migraine and magnesium: eleven neglected connections.
\newblock \emph{Perspectives in biology and medicine}, 31\penalty0
  (4):\penalty0 526--557, 1988.

\bibitem[Wang et~al.(2024)Wang, Downey, Ji, and Hope]{wang-etal-2024-scimon}
Wang, Q., Downey, D., Ji, H., and Hope, T.
\newblock {S}ci{MON}: Scientific inspiration machines optimized for novelty.
\newblock In \emph{Proceedings of the 62nd Annual Meeting of the Association
  for Computational Linguistics (Volume 1: Long Papers)}, pp.\  279--299,
  Bangkok, Thailand, August 2024. Association for Computational Linguistics.
\newblock \doi{10.18653/v1/2024.acl-long.18}.

\bibitem[Weng et~al.(2025)Weng, Zhu, Bao, Zhang, Wang, Zhang, and
  Yang]{weng2025cycleresearcher}
Weng, Y., Zhu, M., Bao, G., Zhang, H., Wang, J., Zhang, Y., and Yang, L.
\newblock Cycleresearcher: Improving automated research via automated review.
\newblock 2025.

\bibitem[Xiong et~al.(2024)Xiong, Xie, Shariatmadari, Guo, Bekiranov, and
  Zhang]{xiong2024improving}
Xiong, G., Xie, E., Shariatmadari, A.~H., Guo, S., Bekiranov, S., and Zhang, A.
\newblock Improving scientific hypothesis generation with knowledge grounded
  large language models.
\newblock \emph{arXiv preprint arXiv:2411.02382}, 2024.

\bibitem[Yang et~al.(2024)Yang, Du, Li, Zheng, Poria, and
  Cambria]{yang2024large}
Yang, Z., Du, X., Li, J., Zheng, J., Poria, S., and Cambria, E.
\newblock Large language models for automated open-domain scientific hypotheses
  discovery.
\newblock In \emph{Findings of the Association for Computational Linguistics:
  ACL 2024}, pp.\  13545--13565, 2024.

\bibitem[Yang et~al.(2026)Yang, Liu, Gao, Liu, Li, Xie, Bing, Ouyang, Cambria,
  and Zhou]{yang2026moose}
Yang, Z., Liu, W., Gao, B., Liu, Y., Li, W., Xie, T., Bing, L., Ouyang, W.,
  Cambria, E., and Zhou, D.
\newblock Moose-chem2: Exploring llm limits in fine-grained scientific
  hypothesis discovery via hierarchical search.
\newblock \emph{Advances in Neural Information Processing Systems},
  38:\penalty0 89045--89076, 2026.

\bibitem[Yuan et~al.(2025)Yuan, Yan, Zhang, Chen, Shi, Ouyang, Qiao, Bai, and
  Zhou]{yuan2025dolphin}
Yuan, J., Yan, X., Zhang, B., Chen, T., Shi, B., Ouyang, W., Qiao, Y., Bai, L.,
  and Zhou, B.
\newblock Dolphin: moving towards closed-loop auto-research through thinking,
  practice, and feedback.
\newblock In \emph{Proceedings of the 63rd Annual Meeting of the Association
  for Computational Linguistics (Volume 1: Long Papers)}, pp.\  21768--21789,
  2025.

\bibitem[Zhou et~al.(2024)Zhou, Liu, Srivastava, Mei, and
  Tan]{zhou-etal-2024-hypothesis}
Zhou, Y., Liu, H., Srivastava, T., Mei, H., and Tan, C.
\newblock Hypothesis generation with large language models.
\newblock In Peled-Cohen, L., Calderon, N., Lissak, S., and Reichart, R.
  (eds.), \emph{Proceedings of the 1st Workshop on NLP for Science
  (NLP4Science)}, pp.\  117--139, Miami, FL, USA, November 2024. Association
  for Computational Linguistics.
\newblock \doi{10.18653/v1/2024.nlp4science-1.10}.

\end{thebibliography}
